\documentclass{article}

\usepackage{arxiv}

\usepackage[utf8]{inputenc} % allow utf-8 input
\usepackage[T1]{fontenc}    % use 8-bit T1 fonts
\usepackage{hyperref}       % hyperlinks
\usepackage{url}            % simple URL typesetting
\usepackage{booktabs}       % professional-quality tables
\usepackage{nicefrac}       % compact symbols for 1/2, etc.
\usepackage{microtype}      % microtypography
\usepackage{natbib}
\usepackage{doi}

\usepackage{amsmath,amssymb,amsfonts}
\usepackage{algorithmic}
\usepackage{graphicx}
\usepackage{textcomp}
\usepackage{xcolor}
\usepackage{multirow}
\usepackage[table,xcdraw]{xcolor}
\usepackage[table,xcdraw]{xcolor}
\usepackage{array}
\usepackage{graphicx} % For \resizebox
\usepackage{amsmath}
\usepackage{subcaption}
\usepackage{balance}
\usepackage{enumitem}
\usepackage{cleveref}\usepackage{longtable}
\usepackage{xurl}
\usepackage{fancyhdr}
\def\bng{\bngx}

\font\bngx=bang10

\def\*#1*#2{o\null{#2}{#1}}

\def\sh#1{\setbox0=\hbox{#1}%
     \kern-.02em\copy0\kern-\wd0
     \kern.04em\copy0\kern-\wd0
     \kern-.02em\raise.0433em\box0 }

\begin{document}

\thispagestyle{empty}
{\Huge \textbf{IEEE Copyright Notice}}\\[1em]
{\large
\noindent
\textcopyright~2026 IEEE. Personal use of this material is permitted. Permission from IEEE must be obtained for all other uses, in any current or future media, including reprinting/republishing this material for advertising or promotional purposes, creating new collective works, for resale or redistribution to servers or lists, or reuse of any copyrighted component of this work in other works.

% DOI: \href{https://doi.org/10.1109/ICCIT64611.2024.11022605}{10.1109/ICCIT64611.2024.11022605}
}

\newpage

\fancypagestyle{titlepage}{
  \fancyhf{}
  \fancyhead[C]{\footnotesize This work has been accepted for publication in IEEE 3rd INTERNATIONAL CONFERENCE ON COMPUTING, APPLICATIONS AND SYSTEMS (COMPAS 2026)\\
  % The final published version is available via IEEE Xplore.\\
  % DOI: \href{https://doi.org/10.1109/ICCIT64611.2024.11022605}{10.1109/ICCIT64611.2024.11022605}
  }
  \renewcommand{\headrulewidth}{0pt}
}

\title{Cross-Dialect NER for Bangla Regional Dialects Using Leave-One-Dialect-Out Cross-Validation and Explainable AI}

%\date{September 9, 1985}	% Here you can change the date presented in the paper title
%\date{} 					% Or removing it

\author{ \href{https://orcid.org/0009-0001-8533-651X}{\includegraphics[scale=0.06]{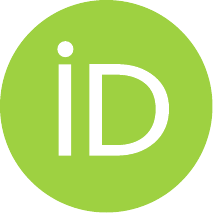}\hspace{1mm}Shamim Rahim Refat} \\
	Department of Computer Science and Engineering\\
	Ahsanullah University of Science and Technology\\
	Dhaka, Bangladesh \\
	\texttt{n.a.refat2000@gmail.com} \\
	%% examples of more authors
	\And
	\href{https://orcid.org/0009-0001-1615-2071}{\includegraphics[scale=0.06]{orcid.pdf}\hspace{1mm}Faika Fairuj Preotee} \\
	Department of Computer Science and Engineering\\
	Southeast University\\
	Dhaka, Bangladesh \\
	\texttt{faikafairuj2001@gmail.com} \\
    \And
	\href{https://orcid.org/0009-0007-6758-0758}{\includegraphics[scale=0.06]{orcid.pdf}\hspace{1mm}Shuvashis Sarker} \\
	Department of Computer Science\\
	Technische Universität Dresden\\
	Dresden, Germany \\
	\texttt{shuvashisofficial@gmail.com} \\
    \And
	\href{https://orcid.org/0009-0008-8433-0264}{\includegraphics[scale=0.06]{orcid.pdf}\hspace{1mm}Shifat Islam} \\
	Department of Computer Science and Engineering\\
	Bangladesh University of Engineering and Technology\\
	Dhaka, Bangladesh \\
	\texttt{shifat.islam.buet@gmail.com} \\
    \And
	\href{https://orcid.org/0009-0003-5892-5786}{\includegraphics[scale=0.06]{orcid.pdf}\hspace{1mm}Bidyarthi Paul} \\
	Department of Computer Science and Engineering\\
	Southeast University\\
	Dhaka, Bangladesh \\
	\texttt{bidyarthipaul01@gmail.com} \\
    \And
	\href{https://orcid.org/0009-0003-5779-1026}{\includegraphics[scale=0.06]{orcid.pdf}\hspace{1mm}Mohammad Ashraful Hoque} \\
	Department of Computer Science and Engineering\\
	Southeast University\\
	Dhaka, Bangladesh \\
	\texttt{ashraful@seu.edu.bd} \\
	%% \AND
	%% Coauthor \\
	%% Affiliation \\
	%% Address \\
	%% \texttt{email} \\
	%% \And
	%% Coauthor \\
	%% Affiliation \\
	%% Address \\
	%% \texttt{email} \\
	%% \And
	%% Coauthor \\
	%% Affiliation \\
	%% Address \\
	%% \texttt{email} \\
}

% Uncomment to remove the date
%\date{}

% Uncomment to override  the `A preprint' in the header
%\renewcommand{\headeright}{Technical Report}
%\renewcommand{\undertitle}{Technical Report}
\renewcommand{\shorttitle}{}

%%% Add PDF metadata to help others organize their library
%%% Once the PDF is generated, you can check the metadata with
%%% $ pdfinfo template.pdf
% \hypersetup{
% pdftitle={Cross-Dialect NER for Bangla Regional Dialects Using Leave-One-Dialect-Out Cross-Validation and Explainable AI},
% pdfsubject={q-bio.NC, q-bio.QM},
% pdfauthor={Shamim Rahim Refat, Faika Fairuj Preotee, Shuvashis Sarker, Shifat Islam, Bidyarthi Paul, Mohammad Ashraful Hoque},
% pdfkeywords={Cross-Dialect Named Entity Recognition, Leave-One-Dialect-Out Cross-Validation (LODOCV), Transformer Models, Bangla BERT, XLM-RoBERTa, MuRil, Multilingual-E5, Explainable Artificial Intelligence (XAI), Local Interpretable Model-Agnostic Explanations (LIME).},
% }

\maketitle
\thispagestyle{titlepage}

\begin{abstract}
Bangla, the seventh most spoken language in the world, exhibits significant regional dialectal diversity, with dialects such as Barishal, Chattogram, Sylhet, Noakhali, and Mymensingh differing in lexical, morphological, and syntactic characteristics. These variations pose substantial challenges for Named Entity Recognition (NER), limiting the generalization of models trained on Standard Bangla or a single regional dialect. This paper presents a cross-dialect Bangla NER framework using the publicly available ANCHOLIK-NER dataset, comprising 17,405 annotated sentences and 101,817 tokens across five major Bangla regional dialects. A Leave-One-Dialect-Out Cross-Validation (LODOCV) strategy is adopted, training models on four dialects and evaluating on the remaining unseen dialect. Eight pretrained transformer-based models, including BanglaBERT, MuRIL, XLM-RoBERTa, and Multilingual-E5, are evaluated under identical experimental settings. Multilingual-E5 Large achieves the highest F1-score in every fold, peaking at 97.26\% on Mymensingh and reaching its lowest, 82.38\%, on Chattogram, the most challenging target dialect. To improve interpretability, Local Interpretable Model-agnostic Explanations (LIME) are applied to word-level predictions, revealing that the model's decisions are driven primarily by the surface form of the target entity word itself rather than by surrounding sentence context. These findings establish a benchmark for cross-dialect Bangla NER and demonstrate the effectiveness of transformer-based transfer learning for low-resource regional dialects, while highlighting the surface-form dependence of current models as a direction for future work.
\end{abstract}

% keywords can be removed

\section{Introduction}
Bangla is the seventh most spoken language in the world, with more than 270 million speakers across Bangladesh and neighboring regions. Despite its widespread usage, Bangla remains a relatively low-resource language in Natural Language Processing (NLP), particularly for dialect-aware applications. Named Entity Recognition (NER), a fundamental NLP task that identifies and classifies entities such as persons, locations, organizations, and other semantic categories, has witnessed remarkable progress through transformer-based architectures \cite{devlin2019bert, bhattacharjee2022banglabert}. However, most existing Bangla NER studies focus on standard written Bangla and assume linguistic homogeneity, overlooking the rich dialectal diversity that characterizes the language.
Bangla consists of numerous regional dialects that differ significantly in vocabulary, phonology, morphology, and syntax. People widely use dialects such as Barishaila, Chittagonian, Sylheti, Noakhailla, and Mymensinghia in daily communication, social media, and regional digital platforms. Previous studies on dialectal NER\cite{el2023cross,moussa2023darnercorp,dahou2023dzner,hamad2025konooz} have demonstrated that linguistic variation can substantially affect model performance and transferability across language varieties. While several dialect-specific NER datasets and benchmarks have been introduced for Arabic and other low-resource languages, the problem of cross-dialect Bangla NER remains largely unexplored due to the scarcity of annotated resources and the lack of comprehensive benchmarking studies.
To address this gap, this study investigates cross-dialect Bangla NER using the ANCHOLIK-NER benchmark dataset, comprising 17,405 annotated sentences and 101,817 tokens collected from five major Bangla regional dialects. Unlike conventional NER settings, where models are trained and evaluated on the same language variety, we focus on the transferability of transformer-based models across unseen dialects. The primary contributions of this work are as follows:
\begin{enumerate}[label=\roman*]
    \item To investigate whether transformer-based NER models trained on four Bangla regional dialects can effectively recognize named entities in an unseen fifth dialect.
    
    \item To evaluate the cross-dialect transferability and generalization capability of different transformer architectures across five major Bangla regional dialects.
    
    \item To interpret and validate model predictions using Explainable Artificial Intelligence (XAI) techniques, providing insights into the linguistic features and contextual cues utilized by transformer models during cross-dialect named entity recognition.
\end{enumerate}

Through these contributions, this work aims to advance dialect-aware Bangla NLP by providing a comprehensive evaluation of transformer-based NER models under realistic cross-dialect settings. The findings provide useful information about the generalization capabilities of modern language models and establish strong baselines for future research on low-resource and dialectal named entity recognition. Furthermore, the proposed benchmark facilitates the development of more robust and inclusive Bangla NLP systems capable of handling linguistic diversity across regional language varieties.

\section{Related Work}
\label{sec:Related_Work}

Recent studies have demonstrated the effectiveness of transfer learning for dialectal Named Entity Recognition (NER), particularly in low-resource settings. Elkhbir et al. \cite{el2023cross} proposed a cross-dialect Arabic NER framework and showed that transformer-based models trained on Modern Standard Arabic (MSA) and multiple dialects can effectively transfer knowledge to unseen dialects under zero-shot settings. Similarly, El Mekki et al. \cite{el2022adasl} introduced AdaSL, an unsupervised domain adaptation framework that combines domain-adaptive pretraining, self-training, and distribution alignment, achieving state-of-the-art performance for zero-shot Arabic NER and POS tagging.
Several studies have focused on constructing benchmark datasets for dialectal NER. Moussa and Mourhir\cite{moussa2023darnercorp} introduced DarNERcorp, the first manually annotated Moroccan Arabic (Darija) NER dataset, comprising 65,905 tokens from 5,000 Wikipedia articles annotated using the BIO scheme. AraBERT achieved the best baseline F1-score of 59.48\%, with the highest performance for location entities. However, the study was limited to Moroccan Arabic, lacked cross-dialect evaluation. Dahou and Cheragui\cite{dahou2023dzner} proposed DzNER, a large-scale Algerian Arabic NER dataset containing over 21,000 manually annotated sentences collected from Facebook and YouTube. After preprocessing and IOB2 annotation, AraBERT achieved the best F1-score of 75.41\%. Mekki et al.\cite{mekki2024named} proposed TUNER, a hybrid Tunisian Arabic NER system combining BiLSTM-CRF with handcrafted linguistic rules. Using data from the Tunisian TreeBank, social media, and news sources, the model achieved a macro F1-score of 91.97\%, particularly improving DATE entity recognition. Likewise, Touileb\cite{touileb2022nerdz}introduced NERDz, the first publicly available Algerian dialect NER dataset with parallel NArabizi, Arabic, and code-switched annotations. The dataset contains 1,276 sentences annotated with eight entity types using the IOB2 scheme. DziriBERT and CNN-BiLSTM-CRF (NCRF++) were evaluated, with NCRF++ achieving the best F1-score of 77.46\% on code-switched text. However, the dataset is relatively small, suffers from class imbalance and tokenization issues. Beyond Arabic, Oyewusi et al.\cite{oyewusi2021naijaner} proposed NaijaNER for five Nigerian languages, demonstrating the feasibility of multilingual NER with a best F1-score of 67.13\%.
More recently, Hamad et al.\cite{hamad2025konooz} introduced Konooz, a large-scale multi-domain and multi-dialect Arabic NER corpus containing 777K tokens across 16 dialects and 10 domains. Their cross-dialect and cross-domain experiments revealed performance degradations of up to 30\% and 38\%, respectively, highlighting the impact of linguistic and domain variation on NER systems. Although these studies have significantly advanced dialectal NER, they primarily focus on Arabic and other multilingual settings. Moreover, most existing works emphasize dataset construction or performance evaluation without incorporating model interpretability. To the best of our knowledge, comprehensive cross-dialect Named Entity Recognition for Bangla regional dialects remains largely unexplored. 

\section{Methodology}
\subsection{Dataset}
The proposed framework is evaluated on the \textbf{ANCHOLIK-NER} (A Benchmark Corpus of Bangla Regional Named Entity Recognition)\cite{paul2026ancholik}, a publicly available benchmark dataset\cite{ancholikNER} covering five major Bangla regional dialects: \textit{Barishal, Chittagong, Sylhet, Noakhali,} and \textit{Mymensingh}. The corpus was constructed from publicly available resources, including the \textit{Vashantor}\cite{faria2023vashantor} and \textit{ONUBAD}\cite{sultana2025onubad} corpora, supplemented with manually translated sentences to ensure a balanced representation of all dialects. The final dataset consists of \textbf{17,405} annotated sentences containing approximately \textbf{101,817} word tokens.

The dataset adopts the standard \textbf{BIO} tagging scheme, where the beginning and inside tokens of an entity are labeled as \texttt{B-<TYPE>} and \texttt{I-<TYPE>}, respectively, while non-entity tokens are assigned the \texttt{O} label. It contains \textbf{9 named entity categories}: \texttt{ANI}, \texttt{COL}, \texttt{FOOD}, \texttt{LOC}, \texttt{OBJ}, \texttt{ORG}, \texttt{PER}, \texttt{REL}, and \texttt{ROLE}.

An example of BIO annotation is shown below, where location names are tagged using the \texttt{B-LOC} label.

\begin{quote}
\textbf{Sylhet Dialect:}

{\bng thaI iselT}/B-LOC {\bng Dhakat}/B-LOC {\bng AaIech.}

\medskip

\textbf{Standard Bangla:}

{\bng iselT}/B-LOC {\bng ethek Dhaka}/B-LOC {\bng Eesech.}
\end{quote}

\begin{table}[htbp]
\centering
\caption{Overview of Ancholik-NER Dataset}
\label{tab:data_stat}
\begin{tabular}{|l|l|}
\hline
\textbf{Dataset Attributes} & \textbf{Frequency} \\ \hline
Total Number of sentences   & 17,405             \\ \hline
Total Named Entities        & 11,062             \\ \hline
Total Non-Named Entities    & 90,755             \\ \hline
Sentence Length             & {[}2-10{]}         \\ \hline
Entities                    & 9                 \\ \hline
Tagging Scheme              & BIO                \\ \hline
Number of Tags              & 19                 \\ \hline
\end{tabular}%
\end{table}

\subsection{Leave-One-Dialect-Out Cross-Validation (LODOCV)}
The cross-dialect evaluation was conducted using a Leave-One-Dialect-Out Cross-Validation (LODOCV) strategy. The ANCHOLIK-NER dataset consists of five regional dialects: Barishal, Chattogram, Sylhet, Noakhali, and Mymensingh. For each LODOCV fold, one dialect was designated as the unseen test set, while the remaining four dialects were merged to form the training set. This process was repeated five times, ensuring that each dialect served as the test set exactly once. Such a setup enables the evaluation of a model's ability to generalize to previously unseen dialectal variations. Each fold is labeled using the initials of the four training dialects followed by the held-out test dialect — e.g., BCMN-S denotes training on Barishal, Chittagong, Mymensingh, and Noakhali, tested on Sylhet. The five folds are thus BCMN-S, BCMS-N, BCNS-M, BMNS-C, and CMNS-B, corresponding to Sylhet, Noakhali, Mymensingh, Chittagong, and Barishal as the respective held-out dialects.

\subsection{Data Preprocessing}
For each LODOCV fold, the datasets from four regional dialects were merged to form the training set, while the remaining dialect was reserved as the unseen test set. The BIO-annotated entity labels were converted into numerical label IDs, and the sentences were tokenized using the tokenizer corresponding to each transformer model. As transformer tokenizers may split a word into multiple subword tokens, the entity labels were appropriately aligned with the tokenized sequences to preserve the original annotations. Finally, the processed data were converted into input IDs, attention masks, and aligned label sequences for model fine-tuning.

\subsection{Proposed Framework}

\begin{figure}[h!]
    \centering
    \includegraphics[width=\linewidth]{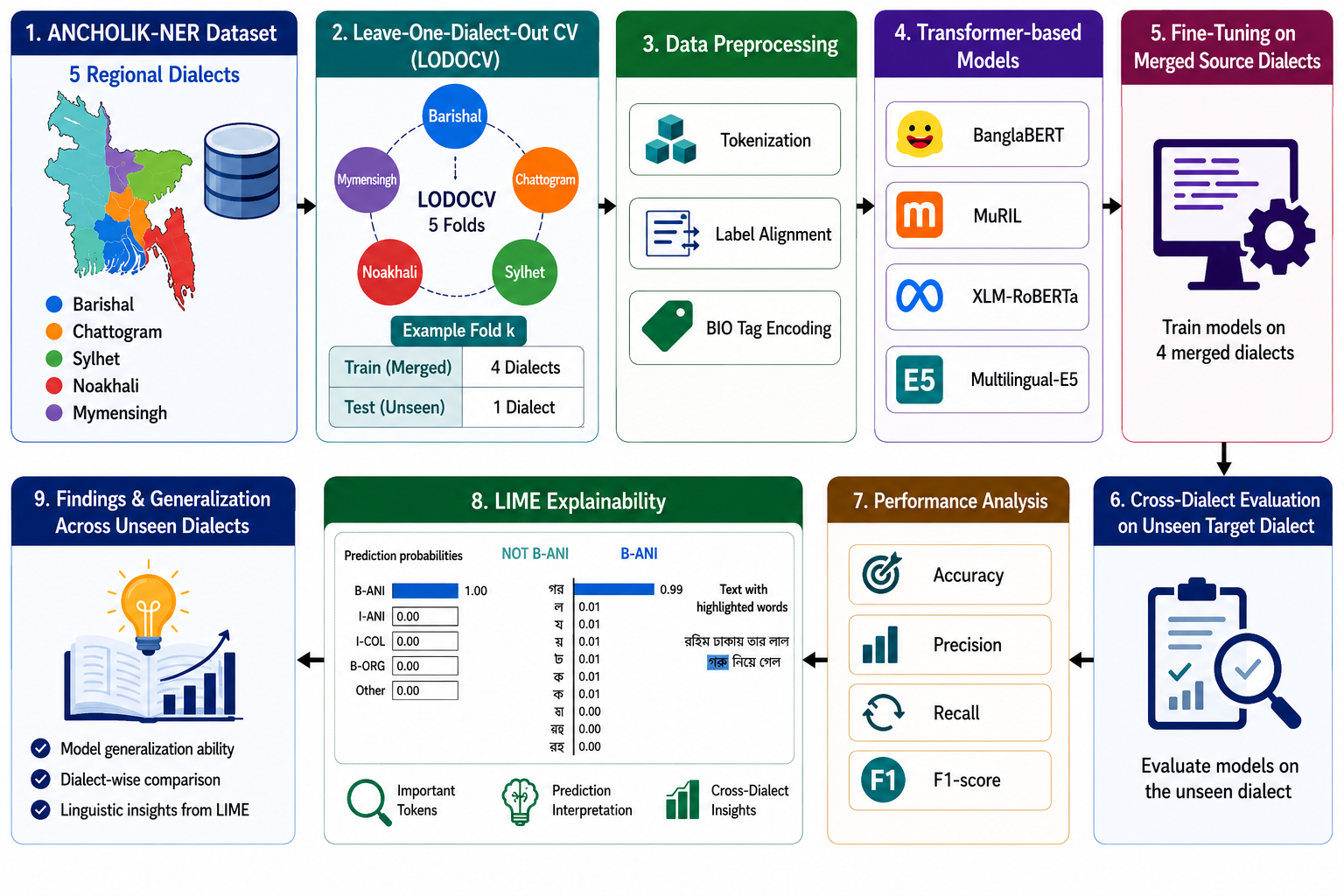}
    \caption{End-to-end Pipeline of the Proposed Cross-dialect Bangla NER Framework.}
    \label{methodology}
\end{figure}

The proposed cross-dialect Named Entity Recognition (NER) framework utilizes the ANCHOLIK-NER dataset, which consists of annotated data from five Bangla regional dialects. Initially, the dataset undergoes tokenization and label alignment to ensure compatibility with transformer-based models. Subsequently, multiple multilingual pretrained language models, including Bangla BERT Base, Bangla BERT Large, XLM-RoBERTa Base, XLM-RoBERTa Large, MuRIL Base Cased, MuRIL Large Cased, Multilingual E5 Base, and Multilingual E5 Large, are fine-tuned for the NER task. To evaluate the cross-dialect generalization capability of these models, a Leave-One-Dialect-Out Cross-Validation (LODOCV) strategy is employed, where four dialects are used for training and the remaining unseen dialect is used for testing. This process is repeated for all five dialects, ensuring that each dialect serves as the test set exactly once. Finally, the models are evaluated using standard NER performance metrics to analyze their effectiveness in recognizing named entities across unseen Bangla regional dialects. Figure \ref{methodology} illustrates the overall workflow.

\subsection{Pre-Trained language Models on Bengali}\label{AA}
To investigate the effectiveness of transformer-based architectures for cross-dialect Bangla Named Entity Recognition (NER), four state-of-the-art pretrained language models were selected: Bangla BERT, MuRIL, XLM-RoBERTa, and Multilingual-E5. These models were chosen due to their strong multilingual capabilities, extensive pretraining corpora, and proven performance on various natural language processing tasks.

\begin{itemize}

\item \textbf{Bangla BERT:} Traditional language models often struggle to capture the subtle linguistic characteristics and contextual nuances of Bangla, particularly across diverse regional dialects. To address this challenge, BanglaBERT\cite{bhattacharjee2022banglabert} is pretrained on a large-scale Bangla corpus, enabling it to learn rich contextual and semantic representations of the language. It's language-specific pretraining enhances the model's ability to accurately recognize and classify entities within dialectal text, thereby improving the robustness and overall performance of Bangla NER systems.

\item \textbf{MuRIL:} MuRIL (Multilingual Representations for Indian Languages)\cite{khanuja2021muril} is a multilingual transformer model developed by Google for Indian languages. It leverages multilingual and translated text during pretraining to improve cross-lingual and low-resource language understanding.

\item \textbf{XLM-RoBERTa:} XLM-RoBERTa\cite{conneau2020unsupervised} is a large-scale multilingual transformer model trained on data from 100 languages. Built upon the RoBERTa architecture, XLM-RoBERTa learns robust multilingual contextual representations from extensive web-scale corpora. The model has demonstrated strong performance in cross-lingual transfer learning and sequence labeling tasks, making it a suitable candidate for evaluating cross-dialect generalization in Bangla NER .

\item \textbf{Multilingual-E5:} Multilingual-E5\cite{wang2024multilingual} is a multilingual embedding model derived from the XLM-RoBERTa architecture and optimized using contrastive learning objectives. Unlike conventional language models, Multilingual-E5 is specifically designed to generate semantically rich text representations across multiple languages. Its enhanced embedding quality and multilingual transfer capabilities enable effective capture of contextual and semantic information, which is beneficial for recognizing named entities in unseen dialectal variations.

\end{itemize}

\begin{table}[htbp]
    \centering
    \caption{Leave-One-Dialect-Out Evaluation Results of Transformer-Based Models}
    \label{tab:placeholder}
    \begin{tabular}{llp{0.07\textwidth}p{0.07\textwidth}p{0.07\textwidth}p{0.07\textwidth}p{0.07\textwidth}p{0.07\textwidth}p{0.07\textwidth}p{0.07\textwidth}} \hline
       Data  &  Metric  &   Bangla BERT Base    &   Bangla BERT Large   &   XLM-RoBERTa Base    &   XLM-RoBERTa Large   &  MuRIL Base Cased    &   MuRIL Large Cased   & Multi-lingual E5 Base  & Multi-lingual E5 Large \\ \hline \hline
        \multirow{4}{*}{BCMN-S} & Precision & 0.89164 & 0.86427 & 0.86541 & 0.83743 & 0.80711 & 0.76157 & 0.85534 & \textbf{0.90214} \\ 
            &   Recall  & 0.83913 & 0.82806 & 0.88710 & 0.83687 & 0.81335 & 0.72995 & 0.90550 & \textbf{0.90940} \\ 
            &   F1-Score    & 0.86459 & 0.84578 & 0.87612 & 0.83715 & 0.81022 & 0.74543 & 0.87971 & \textbf{0.90576} \\ 
            &   Accuracy    & 0.97736 & 0.97577 & 0.97500 & 0.96868 & 0.97017 & 0.96938 & 0.97556 & \textbf{0.98109} \\ \hline
        \multirow{4}{*}{BCMS-N} & Precision & 0.83076 & 0.86450 & 0.90587 & 0.91110 & 0.86881 & 0.93158 & 0.91049 & \textbf{0.91878} \\ 
            &   Recall  & 0.79464 & 0.82725 & 0.87599 & 0.85400 & 0.81637 & 0.86214 & 0.88347 & \textbf{0.88250} \\ 
            &   F1-Score    & 0.81230 & 0.84547 & 0.89068 & 0.88163 & 0.84177 & 0.89551 & 0.89677 & \textbf{0.90028} \\ 
            &   Accuracy    & 0.96966 & 0.97513 & 0.97799 & 0.97623 & 0.97448 & 0.97895 & 0.97876 & \textbf{0.98239} \\ \hline
        \multirow{4}{*}{BCNS-M} & Precision & 0.92797 & 0.94123 & 0.93276 & 0.96175 & 0.92391 & 0.96023 & 0.96028 & \textbf{0.96417} \\ 
            &   Recall  & 0.94682 & 0.95547 & 0.94196 & 0.97425 & 0.94540 & 0.97641 & 0.96969 & \textbf{0.98108} \\ 
            &   F1-Score    & 0.93730 & 0.94829 & 0.93734 & 0.96796 & 0.93453 & 0.96825 & 0.96496 & \textbf{0.97255} \\ 
            &   Accuracy    & 0.98959 & 0.99121 & 0.98826 & 0.99254 & 0.98952 & 0.99454 & 0.99249 & \textbf{0.99387} \\ \hline
        \multirow{4}{*}{BMNS-C} & Precision & 0.77196 & 0.80186 & 0.77482 & 0.80770 & 0.73890 & 0.85667 & 0.79519 & \textbf{0.82904} \\ 
            &   Recall  & 0.68718 & 0.70180 & 0.76949 & 0.79575 & 0.73415 & 0.79112 & 0.79304 & \textbf{0.81867} \\ 
            &   F1-Score    & 0.72711 & 0.74850 & 0.77214 & 0.80168 & 0.73652 & 0.82259 & 0.79411 & \textbf{0.82383} \\ 
            &   Accuracy    & 0.95289 & 0.95673 & 0.95413 & 0.95945 & 0.95154 & 0.96768 & 0.95812 & \textbf{0.96489} \\ \hline
        \multirow{4}{*}{CMNS-B} & Precision & 0.87970 & 0.88874 & 0.88802 & 0.91979 & 0.87551 & 0.93494 & 0.91633 & \textbf{0.93771} \\ 
            &   Recall  & 0.82250 & 0.76719 & 0.85006 & 0.88282 & 0.82237 & 0.88932 & 0.87814 & \textbf{0.90919} \\ 
            &   F1-Score    & 0.85014 & 0.82351 & 0.86863 & 0.90092 & 0.84811 & 0.91156 & 0.89683 & \textbf{0.92323} \\ 
            &   Accuracy    & 0.97510 & 0.97008 & 0.97387 & 0.97946 & 0.97480 & 0.98486 & 0.97957 & \textbf{0.98398} \\ \hline
    \end{tabular}
\end{table}

\section{Result Analysis}
To evaluate the cross-dialect generalization capability of the transformer-based models, we employed a Leave-One-Dialect-Out Cross-Validation (LODOCV) strategy on the ANCHOLIK-NER dataset. In each fold, four regional dialects were merged to form the training set, while the remaining dialect was used as the unseen test set. All models were fine-tuned under identical hyperparameter settings: batch size = 16, learning rate = 3e-5, 10 training epochs, and weight decay = 0.01. Multiple hyperparameter configurations and their combinations were evaluated to determine the optimal settings. Based on the comparative performance analysis, the best-performing hyperparameter combination was selected for the final model training and evaluation. The evaluation was conducted using Precision, Recall, F1-score, and Accuracy.

Table \ref{tab:placeholder} summarizes the performance of eight transformer-based models across the five LODOCV folds. Overall, the Multilingual-E5 Large model consistently achieved the best performance across all evaluation metrics, demonstrating superior cross-dialect generalization. It obtained the highest F1-scores of 90.58\%, 90.03\%, 97.26\%, 82.38\%, and 92.32\% on the BCMN-S, BCMS-N, BCNS-M, BMNS-C, and CMNS-B folds, respectively. Similarly, it achieved the highest accuracies ranging from 96.49\% to 99.39\%, indicating robust token-level classification performance across unseen dialects.

Among the baseline models, XLM-RoBERTa Large and MuRIL Large Cased consistently outperformed their respective base variants, suggesting that increased model capacity contributes to improved dialect generalization. Bangla BERT Large also generally improved upon Bangla BERT Base, although the performance gains were comparatively modest. These observations indicate that larger multilingual transformer models are better equipped to capture lexical and contextual variations across regional Bangla dialects.

Performance varied across the five evaluation folds, reflecting differences in linguistic similarity among the regional dialects. The BCNS-M fold achieved the highest performance across all models, with Multilingual-E5 Large obtaining an F1-score of 97.26\%. One possible explanation is that the Mymensingh dialect is more closely aligned with Standard Bangla than the other regional dialects. Because transformer models are pretrained predominantly on Standard Bangla and similar text, the learned representations may transfer more effectively to the Mymensingh dialect. Conversely, the BMNS-C fold produced the lowest overall performance, with the best F1-score of 82.38\%, suggesting that the Chattogram dialect contains more distinctive lexical and phonological characteristics, which pose greater challenges for cross-dialect transfer.

The consistent improvement achieved by Multilingual-E5 Large across all folds demonstrates its stronger capability to learn generalized contextual representations that transfer effectively to unseen regional dialects. Furthermore, the relatively high accuracies observed across all folds suggest that transformer-based language models can effectively leverage shared semantic and syntactic patterns among Bangla regional dialects while remaining robust to dialect-specific lexical variations.

Overall, the experimental results demonstrate that multilingual transformer models are highly effective for cross-dialect Bangla Named Entity Recognition, with Multilingual-E5 Large providing the best balance between precision, recall, and generalization across unseen regional dialects.

\begin{figure*}[htbp]
    \centering
    \begin{subfigure}[b]{\linewidth}
        \centering
        \includegraphics[width=\linewidth]{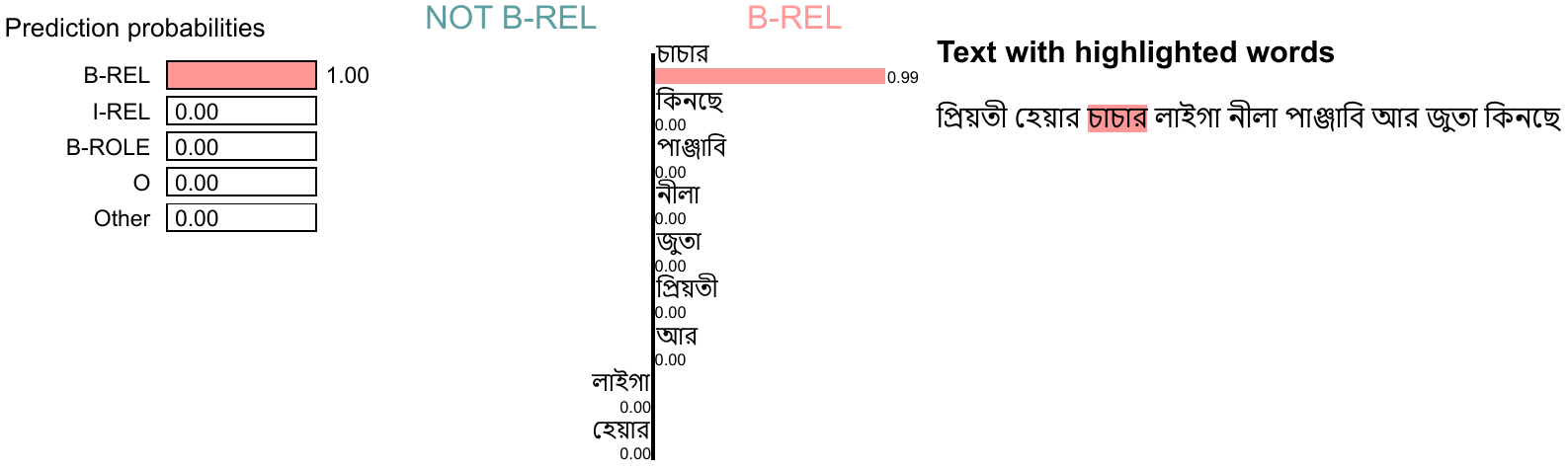} % Replace with your image file
        \caption{}
        \label{fig:subfig1}
    \end{subfigure}
    \begin{subfigure}[b]{\linewidth}
        \centering
        \includegraphics[width=\linewidth]{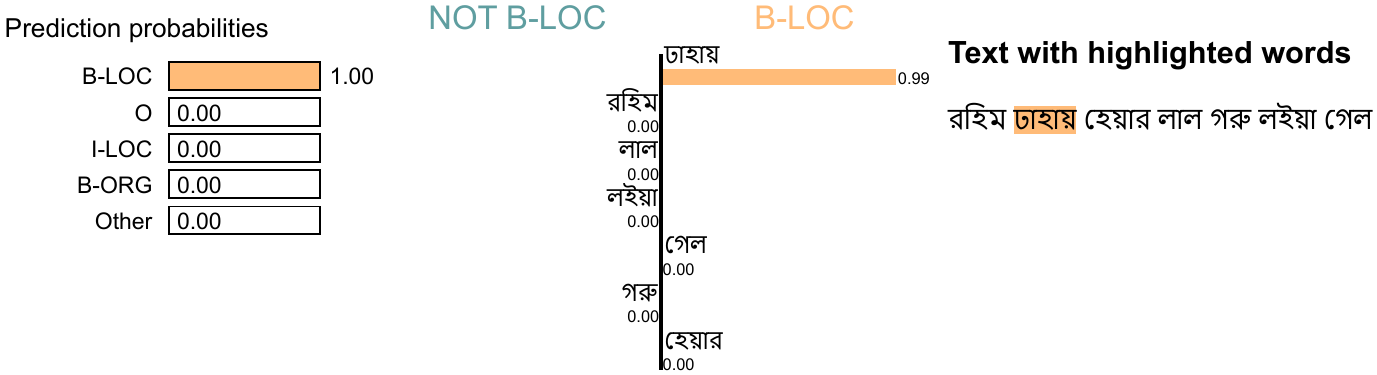} % Replace with your image file
        \caption{}
        \label{fig:subfig2}
    \end{subfigure}
    \caption{LIME-based Interpretation of the Model's Prediction for a Bangla Dialect Token}
    \label{XAI}
\end{figure*}

\section{Explainable AI (XAI) Analysis}
\subsection{Local Interpretable Model-Agnostic Explanations (LIME):}

Local Interpretable Model-Agnostic Explanations (LIME)\cite{tiwari2024hate} is employed to improve the interpretability and transparency of the proposed cross-dialect NER framework. As transformer-based models often operate as black-box systems, understanding the rationale behind their predictions can be challenging. LIME addresses this issue by approximating the behavior of a complex model with a simpler interpretable model around individual predictions. It analyzes the influence of individual words on the model's predictions and identifies the most important tokens responsible for a particular entity classification. In the context of cross-dialect NER, LIME helps reveal the linguistic and contextual cues responsible for entity recognition across different regional dialects. Furthermore, it facilitates error analysis, increases trust in model predictions, and provides valuable insights into the decision-making process of the trained models, thereby enhancing the overall interpretability of the proposed framework.

\subsection{LIME-based Explainability Analysis}

To understand how the model makes its predictions, we employed Local Interpretable Model-agnostic Explanations (LIME) \cite{tiwari2024hate, ribeiro2016should} to analyze individual word-level predictions. Figure~\ref{XAI} presents LIME explanations for two representative Barishal dialect test sentences. The models were trained using the Leave-One-Dialect-Out Cross-Validation (LODOCV) setting, where Barishal was excluded from training and the remaining four dialects (Chittagong, Mymensingh, Noakhali, and Sylhet) were used to learn transferable entity representations.

Figure~\ref{XAI}(a) shows the result for the word ``{\bng cacar}'' (uncle's) in the sentence ``{\bng ipRyit eHyar cacar laIga niila pan/jaib Aar juta iknech}'' ("Priyoti bought a blue panjabi and shoes for her uncle"). The model correctly labels this word as B-REL (start of a relation entity) with full confidence (1.00). LIME shows that almost all of this confidence (0.99) comes from the word ``{\bng cacar}'' itself. Every other word in the sentence — even the ones right next to it — contributes almost nothing (0.00).

Figure~\ref{XAI}(b) shows a similar result for the word ``{\bng DhaHay}'' (to Dhaka, in the Barishal dialect) in the sentence ``{\bng riHm DhaHay eHyar lal gru lIya egl.}'' ("Rahim took his red cow to Dhaka"). The model correctly labels this as B-LOC (start of a location entity), again with full confidence. As before, LIME shows that the prediction depends almost entirely on the target word ``{\bng DhaHay}'' (0.99), while the rest of the sentence has no real effect.

In both examples, the model's decision comes almost entirely from the entity word itself, not from the words around it. This suggests that the model recognizes entities mainly by their form — how the word looks — rather than by using context clues from the rest of the sentence. This pattern was the same for both a relation word and a location word, suggesting the model's ability to work across dialects comes from directly recognizing dialect-specific words, not from understanding sentence-level context.

\section{Discussion and Conclusion}
The experimental results demonstrate that transformer-based models can transfer named entity recognition knowledge across unseen Bangla regional dialects to a substantial degree. Using the proposed Leave-One-Dialect-Out Cross-Validation (LODOCV) framework, models trained on four dialects generalized to a previously unseen fifth dialect, with Multilingual-E5 Large achieving the highest F1-score in every fold — from 82.38\% on the most challenging dialect, Chattogram, to 97.26\% on Mymensingh. This roughly 15-point spread indicates that dialectal variation has a measurable effect on cross-dialect transferability, with Chattogram's more distinctive lexical and phonological characteristics posing the greatest challenge to generalization. It is worth noting that Multilingual-E5 Large's advantage, while consistent on F1-score and Accuracy, does not extend to every individual metric in every fold — MuRIL Large Cased attains marginally higher precision on the Noakhali fold, for instance — suggesting the performance gap between the strongest multilingual models is real but not absolute.

The LIME-based explainability analysis offered a more tempered picture of model behavior than the aggregate metrics alone suggest. Rather than confirming that the models draw on broad linguistic or morphological reasoning, the analysis showed that predictions for both a relation entity and a location entity were driven almost entirely by the target word's own surface form, with negligible contribution from surrounding context. This suggests that the strong cross-dialect F1-scores reported here may reflect the models' ability to recognize dialect-specific lexical forms directly, rather than a deeper contextual or syntactic understanding of dialectal Bangla. This distinction is an important caveat for interpreting the headline results and suggests that reported performance may be sensitive to whether comparable surface forms were present in the training dialects — a question that warrants further investigation, including analysis of lexical overlap between the source corpora underlying different dialect variants.

Overall, this work establishes a benchmark for cross-dialect Bangla NER and shows that transfer learning is a promising approach for low-resource dialectal settings, while also surfacing open questions about what drives that transfer. Future work will extend this framework to additional Bangla dialects and larger datasets, incorporate entity-level and per-class evaluation metrics, quantify lexical overlap across dialect corpora to rule out data leakage as a contributor to the observed performance, and apply word- and subword-level explainability methods more broadly to build a clearer picture of what these models actually learn from dialectal variation.

\bibliographystyle{unsrt}
\bibliography{references}

\end{document}